\documentclass[10pt,twocolumn,letterpaper]{article}

\usepackage[pagenumbers]{arxiv} 

\usepackage{bbm}
\usepackage{booktabs}
\usepackage{colortbl}
\usepackage{float}
\usepackage{dblfloatfix}
\usepackage{multirow}
\usepackage{xcolor}
\newcommand{\CC}[0]{\cellcolor{LightGray}}
\definecolor{LightGray}{gray}{0.9}
\definecolor{DarkRed}{rgb}{0.6, 0, 0}
\definecolor{White}{rgb}{1, 1, 1}

\usepackage[pagebackref,breaklinks,colorlinks]{hyperref}

\title{Towards Cross-Domain Multi-Targeted Adversarial Attacks}

\author{Taïga Gonçalves, Tomo Miyazaki, Shinichiro Omachi\\
Graduate School of Engineering, Tohoku University, Department of Communications Engineering\\
Sendai, Japan\\
{\tt\small goncalves.taiga.teo.t6@dc.tochoku.ac.jp}, {\tt\small tomo@tohoku.ac.jp}, {\tt\small shinichiro.omachi.b5@tohoku.ac.jp}
}

\begin{document}
\maketitle

\begin{abstract}
		Multi-targeted adversarial attacks aim to mislead classifiers toward specific target classes using a single perturbation generator with a conditional input specifying the desired target class. Existing methods face two key limitations: (1) a single generator supports only a limited number of predefined target classes, and (2) it requires access to the victim model's training data to learn target class semantics. This dependency raises data leakage concerns in practical black-box scenarios where the training data is typically private. To address these limitations, we propose a novel Cross-Domain Multi-Targeted Attack (CD-MTA) that can generate perturbations toward arbitrary target classes, even those that do not exist in the attacker's training data. CD-MTA is trained on a single public dataset but can perform targeted attacks on black-box models trained on different datasets with disjoint and unknown class sets. Our method requires only a single example image that visually represents the desired target class, without relying its label, class distribution or pretrained embeddings. We achieve this through a Feature Injection Module (FIM) and class-agnostic objectives which guide the generator to extract transferable, fine-grained features from the target image without inferring class semantics. Experiments on ImageNet and seven additional datasets show that CD-MTA outperforms existing multi-targeted attack methods on unseen target classes in black-box and cross-domain scenarios. The code is available at \url{https://github.com/tgoncalv/CD-MTA}.
\end{abstract}

\section{Introduction}

Deep Neural Networks (DNNs) are widely deployed in safety-critical domains including autonomous driving~\cite{SelfDriving_Ndikumana_1,zhouStealthyEffectivePhysical2024,luCrosstaskTimeawareAdversarial2025} and medical imaging~\cite{UNet,medic_Liu1}. However, adversarial examples, generated by adding subtle noise to the inputs, can significantly disturb their decisions. These perturbations threaten diverse tasks such as image~\cite{AdvAtt_Pandas,AdvAtt_GD-UAP}, face~\cite{zhengRobustPhysicalWorldAttacks2023,huAttentionguidedEvolutionaryAttack2023}, and speech recognition~\cite{ASR_zipformer,kimGeneratingTransferableAdversarial2023}. Thus, discovering and understanding vulnerabilities of DNNs is essential to assess their safety and reliability.

Adversarial attacks can be categorized into white-box and black-box attacks. In white-box attacks, the attacker has full access to the model architecture and parameters, allowing for more effective noise generation. In contrast, black-box attacks assume no prior knowledge of the victim model, making them more challenging. Black-box attacks can be achieved through query-based~\cite{baiQueryEfficientBlackbox2023,mumcuSequentialArchitectureagnosticBlackbox2024,huQueryefficientBlackboxEnsemble2025} or transfer-based~\cite{AdvAtt_TTP,AdvAtt_TTAA,zengStaircaseSignMethod2026} methods. Query-based attacks iteratively refine the perturbation by querying the victim model multiple times. On the other hand, transfer-based attacks generate adversarial examples using a surrogate white-box model, which are then expected to be effecttive against diverse black-box models without additional queries. A popular strategy in transfer-based attacks is the generator-based method~\cite{AdvAtt_DGTA,AdvAtt_GAP,AdvAtt_CD-AP}, which trains a perturbation generator that can transform any input image into an adversarial example with a single forward pass.

Depending on the objective, adversarial attacks can be either untargeted or targeted. Untargeted attacks aim to mislead the victim model into any incorrect prediction, while targeted attacks attempt to redirect it toward a specific target class~\cite{AdvAtt_TTP,AdvAtt_TTAA,AdvAtt_DGTA}. Recent studies highlighted that generator-based targeted attacks can only generate perturbations for a single target class per generator \cite{AdvAtt_OnceMAN,AdvAtt_C-GSP}. This single-targeted approach is impractical as it requires heavy computational resources to train separate generator for each target class. To address this limitation, multi-targeted attacks have been proposed, which train a single generator with a conditional input to control the target class~\cite{AdvAtt_ESMA,AdvAtt_CGNC,AdvAtt_GAKer}.

\begin{figure*}[!t]
	\centering
	\includegraphics[width=0.95\linewidth]{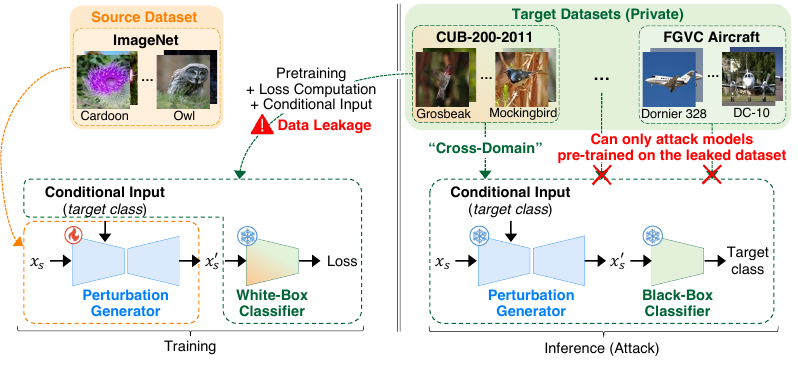}
	\caption{
		\textbf{Illustration of the data leakage problem in existing cross-domain targeted attacks}~\cite{AdvAtt_TTP,AdvAtt_TTAA,AdvAtt_CD-AP,AdvAtt_CGNC,AdvAtt_FDAbis}. These methods are traditionally considered cross-domain as long as the input images $x_s$ of the perturbation generator are from a different dataset between training and evaluation. However, existing methods still use private information from the target dataset during training---such as ground-truth labels and pretrained white-box classifiers---leading to unintended data leakage. As a result, these attacks are bound to a specific source-target dataset pair and cannot be directly applied to other datasets without retraining. This contradicts the initial goal of cross-domain attacks, which is to generalize to unknown datasets. In contrast, our method eliminates data leakage by removing all components and losses that depend on the target dataset.
	}
	\label{fig:data_leakage}
\end{figure*}

Most prior works on black-box targeted attacks overlook a crucial practical consideration: the victim's training dataset is generally private and inaccessible to the attacker. Some studies highlighted this issue \cite{AdvAtt_CD-AP,AdvAtt_LTP} and proposed cross-domain attacks, but mostly in untargeted settings. The targeted version is far more difficult because the attacker cannot learn the target class distribution during training if the corresponding dataset is unknown. As a result, achieving cross-domain targeted attacks remains an open challenge, and only a few works attempted to perform such attacks \cite{AdvAtt_TTP,AdvAtt_TTAA,AdvAtt_CGNC}. The common setup in these works is to feed images from a \textit{source dataset} to their perturbation generator during training, while they use a different \textit{target dataset} during inference. However, as illustrated in Fig.~\ref{fig:data_leakage}, only the generator part adheres to the cross-domain setting. Other components of the attack pipeline implicitly use private information from the target dataset during training, such as ground-truth labels and pretrained classifiers. This data leakage problem undermines the validity of the cross-domain evaluation protocol in existing targeted attacks. Since these methods implicitly use information from a specific target dataset during training, they are effective only for that dataset and require retraining for each new dataset. In contrast, untargeted cross-domain attacks have demonstrated the ability to generalize across multiple unknown datasets without retraining~\cite{AdvAtt_BIA,fangOnePerturbationEnough2025,pengGenerativeAttackComplex2026}


Our work achieves cross-domain targeted attacks without any usage of the target dataset during training. The main challenge lies in the absence of prior knowledge about the target classes. Since label spaces differ across datasets, a target class encountered during inference might not exist in the attacker's source dataset (Fig~\ref{fig:data_leakage}). To address this, we propose Cross-Domain Multi-Targeted Attack (CD-MTA), which uses an image-based conditional input that represent the target class. This input does not require explicit class information, but rather an arbitrary representation of the desired class. Then, we introduce class-agnostic objectives to generate adversarial examples that are spatially aligned with the conditional input in the feature space. CD-MTA does not rely on any class-specific knowledge: we deliberately avoid using class labels, dataset-dependent distributions, or pretrained embeddings, replacing them with abstract feature representations. Our experiments demonstrate that existing methods, which rely on class-specific information, tend to overfit to the target classes used during training, while CD-MTA generalizes more effectively to unseen classes. 

In summary, our main contributions are as follows:

\begin{itemize}
	\item We identify a critical data leakage issue in prior cross-domain targeted attacks, where information from the target dataset is implicitely used during training. To address this, we introduce the \textbf{C}ross-\textbf{D}omain \textbf{M}ulti-\textbf{T}argeted \textbf{A}ttack (\textbf{CD-MTA}), a method capable of performing targeted attacks against any black-box classifiers without requiring prior knowledge of the target classes or datasets.

	\item We propose a new framework that replaces class-specific semantic information with fine-grained features extracted from a conditional target image. To this end, we design a Feature Injection Module (FIM) and class-agnostic feature objectives that encourage the generator to align adversarial examples with the target image in the feature space, without relying on class labels or embeddings.

	\item Extensive experiments demonstrate that CD-MTA achieves state-of-the-art performance on ImageNet across nine black-box classifiers with unseen target classes, following the benchmark used by prior multi-targeted attacks. Moreover, we show that CD-MTA generalizes to seven completely unknown datasets in cross-domain settings without any retraining. Our analysis further reveals that existing methods often overfit to the target classes seen during training, which explains their limited effectiveness on unseen classes.
\end{itemize}

\section{Related Work}

\subsection{Adversarial Attacks}


Pioneering works on adversarial attacks focused on instance-specific methods, where a unique perturbation is generated for each input image. The Fast Gradient Sign Method (FGSM)~\cite{AdvAtt_Pandas} and its iterative variants~\cite{AdvAtt_IFGSM,AdvAtt_PGD} are among the most popular techniques in this category. Subsequent works enhanced attack effectiveness by optimizing various objectives, such as the Carlini \& Wagner (CW) loss~\cite{AdvAtt_CW}, momentum~\cite{AdvAtt_MIM}, utilizing diverse inputs~\cite{AdvAtt_DIM}, and generating translation-invariant perturbations~\cite{AdvAtt_TI}.

Later, instance-agnostic attacks~\cite{AdvAtt_GD-UAP,AdvAtt_UAP} proposed to train a unique noise that can be applied to any image. These noises, known as Universal Adversarial Perturbations (UAPs)~\cite{AdvAtt_selfU,AdvAtt_TRM-UAP,zhangInterpretingVulnerabilitiesMultiinstance2023}, inspired the development of generator-based attacks~\cite{AdvAtt_GAP,AdvAtt_cGAN}, which train a perturbation generator that is capable to produce a noise specific to each image with a single forward pass. This approach greatly improved computational efficiency, noise diversity, and attack transferability~\cite{AdvAtt_TTP,AdvAtt_DGTA,AdvAtt_GAP}. However, in targeted settings, most of these methods can only generate noises for a unique target class. This single-targeted approach is poorly efficient in practice due to the cost of retraining the attack for each target class.

\subsection{Multi-Targeted Attacks}

Recent studies have explored multi-targeted attacks by adding a conditional input to generator-based methods. OnceMAN~\cite{AdvAtt_OnceMAN} and C-GSP~\cite{AdvAtt_C-GSP} condition the generator on class labels, while ESMA~\cite{AdvAtt_ESMA} enhances this approach by focusing the training on High-Sample Density Region (HSDR). However, such label-based conditioning can only support a limited number of target classes per generator~\cite{AdvAtt_C-GSP,AdvAtt_CGNC}. To improve scalability, CGNC~\cite{AdvAtt_CGNC} replaces class labels with CLIP text embeddings~\cite{CLIP}, while GAKer~\cite{AdvAtt_GAKer} uses images as conditional inputs. Dual-Flow~\cite{adv_dualflow} further introduces a diffusion-based~\cite{StableDiffusion} multi-targeted attack that also relies on CLIP embeddings as conditional inputs. Despite these advancements, these methods remain overly dependent on class-specific embeddings, such as classifier logits or dense features from the final layers of a white-box model. Our experiments also show that this over-reliance on class semantics leads to overfitting to the classes encountered during training, resulting in poor generalization to novel target clases. Similar to GAKer, our method also adopts image-based conditioning, but we explicitly avoid relying on class semantics, thus eliminating overfitting and enabling class-agnostic targeted attacks.

\subsection{Cross-Domain Attacks}

Recent methods mainly focus on black-box settings, where the internal parameters and architecture of the victim model are unknown~\cite{AdvAtt_TTP,AdvAtt_DGTA,AdvAtt_GAP,AdvAtt_FastFeatureFool}. The cross-domain setting introduces an additional constraint by assuming no access to the original training data. To address this, existing methods use transfer-based techniques~\cite{AdvAtt_BIA,fangOnePerturbationEnough2025,liUCGUniversalCrossDomain2024,yangPromptDrivenContrastiveLearning2024} by attacking intermediate features of a white-box model. This strategy is widely employed because neural networks share similar activation patterns across models, domains, and tasks at intermediate layers~\cite{AdvAtt_LTP}. While effective for untargeted attacks, these methods are fundamentally limited in targeted settings. Specifically, they lack the ability to steer predictions toward a specific class, as they merely disrupt feature representations rather than inducing a controlled, class-specific shift.

Several works have attempted cross-domain targeted attacks~\cite{AdvAtt_TTP,AdvAtt_TTAA,AdvAtt_CD-AP,AdvAtt_CGNC,AdvAtt_FDAbis}, but suffer from data leakage, as illustrated in \cref{fig:data_leakage}. For instance, \text{CD-AP}~\cite{AdvAtt_CD-AP} uses ground-truth labels from the unknown target dataset to compute a cross-entropy loss during training. TTP~\cite{AdvAtt_TTP} and CGNC~\cite{AdvAtt_CGNC} exploit all the images of the target dataset to learn the distribution of the target classes. Furthermore, all these methods implicitly assume access to a surrogate white-box model pretrained on the private target dataset, which is a non-trivial assumption in practical scenarios. As a result, these attacks are tightly coupled to a specific target dataset and cannot generalize effectively to other datasets unless explicitly retrained. In contrast, CD-MTA achieves true cross-domain multi-targeted attacks without any access to the target dataset or its labels during training. A single trained CD-MTA model can generate perturbations toward arbitrary target classes from any unseen dataset

\section{Methodology}

\begin{figure*}[!t] \centering
	\includegraphics[width=1.0\linewidth]{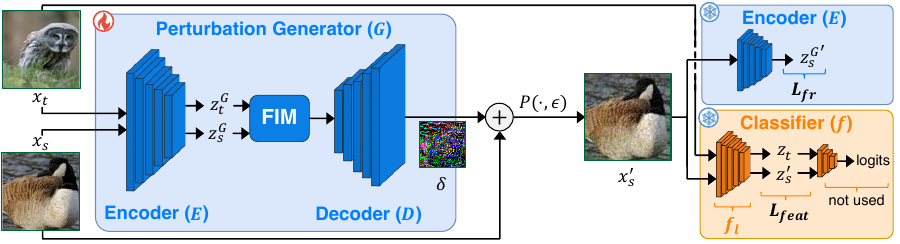}
	\caption{\textbf{Overview of CD-MTA.} The Feature Injection Module (FIM) merges the source and target images and produce a perturbation $\delta$. Then, an adversarial example $x_s'$ is formed using the projection function $P(\cdot,\epsilon)$, as in \cref{eq:perturbed_img_formula}. During training, a white-box classifier extracts the intermediate feature maps and computes the feature loss $L_{feat}$, while the encoder $E$ is reused to compute the feature reconstruction loss $L_{fr}$.}
	\label{fig:framework_overview}
\end{figure*}

\subsection{Problem Formulation}\label{sec:overview}

Let $f: \mathbb{R}^{H \times W \times C} \rightarrow \mathbb{R}^K$ be a classifier that maps an image $x$ of height $H$, width $W$, and $C$ channels to a $K$-dimensional output vector, where $K$ is the number of classes. Given a pair of source and target images $(x_s,x_t)$, our goal is to generate a noise $\delta$ that satisfies the following constraint:

\begin{equation}\label{eq:constraint}
	\delta = \min_{\delta} \ell\left( f(x_s+\delta),f(x_t) \right) \quad \text{s.t.} \quad \left\| \delta \right\|_p \leq \epsilon,
\end{equation}
where $\ell(\cdot,\cdot)$ is a discrepancy loss between the source and target predictions, and $||\cdot||_p$ is typically the $L_\infty$-norm. The constraint $\left\| \delta \right\|_p \leq \epsilon$ ensures that the perturbation is imperceptible to the human eye. The goal is for the pertubed image $x_s + \delta$ to be misclassified as the same class as $x_t$, under a limited perturbation budget $\epsilon$.

\subsection{Preliminary}\label{sec:preliminary}

In this section, we briefly summarize the mechanism of GAKer~\cite{AdvAtt_GAKer}, as it shares several similarities with our proposed method. GAKer is a generator-based multi-targeted attack that takes as input a source image $x_s$ to be perturbed, and a target image $x_t$ representing the desired target class. Unlike label-based conditioning methods, GAKer uses $x_t$ instead of a one-hot vector, enabling generalization to unseen classes at inference time.

To condition the perturbation generator, the target image $x_t$ is embedded using a frozen white-box classifier $f$, while the source image $x_s$ is processed by a trainable encoder. The two embeddings are merged via a Feature Transform Module, and the resulting representation is passed through a decoder to generate the perturbation $\delta$. The adversarial example $x_s'$ is obtained by adding $\delta$ to $x_s$. The model is trained using logit-based losses computed on $\delta$, $x_s'$, and $x_t$, all evaluated via the same classifier $f$.


Despite its advantages, GAKer suffers from two major limitations. First, since $x_t$ is always encoded via $f$, the classifier must be available at inference time and must also be capable of extracting meaningful embeddings for unseen target classes. This can only be satisfied if $f$ has been pretrained on the target dataset. This requirement limits the applicability of GAKer in practical cross-domain scenarios where such pretrained classifier is generally unavailable. Second, GAKer relies on logit-based losses to train the attack. However, these logits inherently encode class-specific semantics and may fail to generalize to novel classes outside the training set, resulting in limited transferability in true cross-domain settings. To overcome these limitations, CD-MTA must introduce a new architecture that does not depend on a pretrained classifier, as well as class-agnostic loss functions that do not rely on class-specific semantics.

\subsection{Architecture of CD-MTA}

We present the architecture of our proposed framework in \cref{fig:framework_overview}. Our perturbation generator $G(\cdot,\cdot)$ receives a source image $x_s$ and a target image $x_t$ as inputs and outputs a perturbation $\delta$. The perturbed image $x_s'$ is then obtained using the following projection function:

\begin{align}\label{eq:perturbed_img_formula}
	x_s' & = P(x_s + \delta, \epsilon),                                                                               \\
	     & = \text{Clip} \Big( \mathcal{W} \cdot \big( x_s + G(x_t, x_s) \big), x_s - \epsilon, x_s + \epsilon \Big),
\end{align}
where $P(\cdot,\epsilon)$ enforces the $L_p$-norm constraint from \cref{eq:constraint} through a clipping operation $\text{Clip}(\cdot)$. The term $\mathcal{W}$ is a differentiable Gaussian kernel used to enhance attack transferability~\cite{AdvAtt_TTP,AdvAtt_C-GSP,AdvAtt_ESMA}.

\begin{figure}[!t] \centering
	\includegraphics[width=0.90\linewidth]{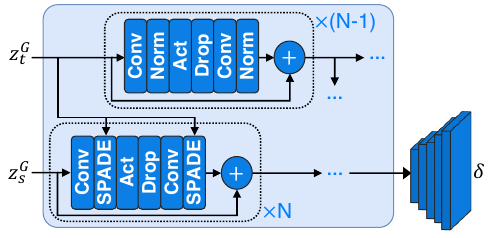}
	\caption[Feature Injection Modules (FIM)]{Feature Injection Module (FIM): the target information is injected into the source using SPADE normalization~\cite{SPADE}.}
	\label{fig:FIM}
\end{figure}

The generator follows a ResNet-like autoencoder architecture inspired by GAKer, but with a crucial distinction. Instead of relying on a pretrained white-box classifier to encode the target image, we process both $x_s$ and $x_t$ using a shared trainable encoder $E$ to obtain their feature maps. These features are then fused using our proposed Feature Injection Module (FIM), before being passed to the decoder to generate the perturbation $\delta$.

The FIM consists of two parallel residual branches to process the source and target features, as shown in \cref{fig:FIM}. The target features are injected into the source features using SPADE normalization~\cite{SPADE}. Specifically, SPADE uses convolutional layers to compute spatially adaptive parameters $\beta$ and $\gamma$ from the target features and applies them to normalize the source features:

\begin{equation}\label{eq:spade_norm}
	z_s^{G} = \gamma \cdot \frac{z_s^G - \mu_s^G}{\sigma_s^G} + \beta,
\end{equation}
where $\mu_s^G$ and $\sigma_s^G$ are channel-wise spatial statistics of the source features $z_s^G$. The advantage of SPADE is the preservation of spatial layouts during normalization, which is crucial for transferring localized patterns.

By employing a shared encoder, we eliminate the need for a pretrained white-box classifier to process the target image $x_t$, thereby improving the practicality of our method in cross-domain scenarios. However, this design introduces a new challenge: the generator must learn to extract class-relevant information from $x_t$ without the guidance from pretrained semantic representations. This issue is specifically addressed by our FIM, which enables the generator to leverage local and structural patterns from $x_t$. This allows the model to focus on fine-grained features that generalize across classes and datasets, rather than depending on dataset-specific class semantics. The effectiveness of this strategy is further reinforced by the loss functions described in the following sections.

\subsection{Class-Agnostic Feature Objective}

Our primary loss is based on intermediate features of the white-box source classifier. Given a classifier $f$, we denote $f_l:~\mathbb{R}^{H \times W \times C} \rightarrow \mathbb{R}^{h_l \times w_l \times c_l}$ as the feature maps of size $h_l \times w_l \times c_l$ extracted from the $l$-th layer. Then, we define the feature loss as:

\begin{align}\label{eq:feature_loss}
	L_{feat} & = 1 - \frac{f_l(x_s') \cdot f_l(x_t)}{\left\| f_l(x_s') \right\| \left\| f_l(x_t) \right\|}.
\end{align}


Existing multi-targeted attacks rely on class-specific supervision, using cross-entropy~\cite{AdvAtt_C-GSP,AdvAtt_OnceMAN,AdvAtt_CGNC,adv_dualflow} or distance-based losses~\cite{AdvAtt_ESMA,AdvAtt_GAKer} computed on the $K$-dimensional logit vectors, where $K$ is the number of classes of the source dataset. This makes them tightly coupled to the structure of the training dataset, limiting their adaptability to new labels from unknown datasets.

Some variants of our loss in \cref{eq:feature_loss} have been explored in untargeted attacks~\cite{AdvAtt_BIA,yangPromptDrivenContrastiveLearning2024,yangFACLAttackFrequencyAwareContrastive2024}, but their application in targeted settings is non-trivial. In fact, a single target sample $x_t$ typically fails to capture the full variability of the target class distribution due to spatial inconsistencies across samples~\cite{AdvAtt_PAA-GAA}. To mitigate this, prior feature-based targeted attacks have relied on spatially invariant representations---either by applying statistical pooling operations (\textit{e.g.}, mean or standard deviation of feature maps)~\cite{AdvAtt_PAA-GAA,AdvAtt_StylePert} or by training binary discriminators~\cite{AdvAtt_TTAA,AdvAtt_FDAbis} to learn the internal feature distribution of the target class. However, these strategies are only viable in single-targeted settings where the target class is known and fixed in advance.

In contrast, our method treats $x_t$ as a conditional input to the generator and preserves its spatial structure through our FIM. This ensures that the resulting adversarial example $x_s'$ is spatially aligned with $x_t$. As a result, we can directly compare the intermediate features of $x_s'$ and $x_t$ using \cref{eq:feature_loss}, without the need to apply any pooling operation that would require multiple samples per target class. This design removes the need for class-specific or dataset-dependent supervision and enables a more scalable and class-agnostic formulation of targeted attacks.


\begin{figure}[!t]
	\centering
	\includegraphics[width=1.00\linewidth]{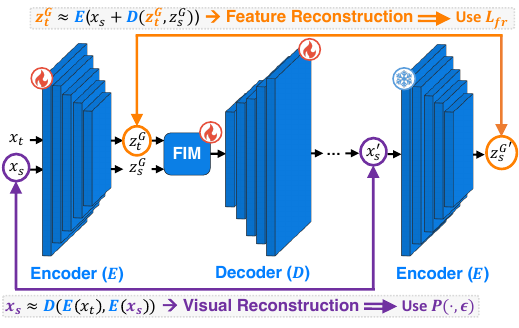}
	\caption[Feature Reconstruction Objective (FRO)]{
		\textbf{Illustration of the Feature Reconstruction Objective (FRO)}. The perturbation generator is trained to reconstruct the source data $x_s$ in the visual space using $P(\cdot, \epsilon)$ as in \cref{eq:perturbed_img_formula}, while simultaneously reconstructing the target data $z_t^G$ in the feature space using the feature reconstruction loss $L_{fr}$ in \cref{eq:FRO}.

	}
	\label{fig:FRO}
\end{figure}

\subsection{Dual Reconstruction Objective}

Traditional autoencoders are designed to reconstruct the input in the visual space. In our case, this corresponds to reconstructing the source image $x_s$ through the generator pipeline, as formalized by:

\begin{equation}\label{eq:visual_reconstruction}
	x_s \approx D\left(E(x_t), E(x_s)\right),
\end{equation}
where $E(\cdot)$ is the encoder and $D(\cdot,\cdot)$ represents the combination of our FIM and the decoder. This reconstruction task is already enforced in all generator-based attacks, including CD-MTA, via the projection function in \cref{eq:perturbed_img_formula}.


However, unlike standard autoencoders that operate on a single image, our model takes two input, $x_s$ and $x_t$, and produces a single output $x_s'$. The visual space reconstruction only applies to the source data. To complement this, we introduce a second reconstruction task that applies to the target data using the feature space, as illustrated in \cref{fig:FRO}. Specifically, we aim to reconstruct the target features $z_t^G$ from the perturbed image $x_s'$, as follows:

\begin{equation}\label{eq:feature_reconstruction}
	z_t^G \approx E\left(P(x_s + D(z_t^G, z_s^G), \epsilon)\right).
\end{equation}

This dual reconstruction mechanism ensures that the adversarial examples simultaneously preserves the source data in the visual space for human perception, and the target data in the feature space for machine perception. To formalize the second objective, we define the Feature Reconstruction Objective (FRO) as:

\begin{equation}\label{eq:FRO}
	L_{fr} = 1 - cos\left(E(x_s'), E(x_t)\right).
\end{equation}

To prevent feature collapse and preserve meaningful gradients for the generator, we freeze the encoder $E$ when backpropagating $L_{fr}$. Our final loss is defined in \cref{eq:total_loss}, where $\lambda_{feat}$ is a hyperparameter:

\begin{equation}\label{eq:total_loss}
	L_{total} = \lambda_{feat} \cdot L_{feat} + L_{fr}.
\end{equation}

\section{Experiments}

\subsection{Experimental Setup}

\textbf{Training Data.} We used ImageNet~\cite{dataset_ImageNet} as the training dataset for all experiments, with 200 classes for training and the remaining 800 classes for evaluation. The experimental setup, including the construction of the training and evaluation dataset, follows the same procedure as GAKer~\cite{AdvAtt_GAKer}. This ensures a fair comparision between our model and other baselines. Specifically, we chose the same 200 classes as GAKer for training. For each class, we used the 325 images with the lowest cross-entropy loss, resulting in a total of 65k training images per epoch.

\begin{table}[!t] \centering
	\caption{Detailed architecture of CD-MTA.}
	\renewcommand\arraystretch{1.3}
	\resizebox{0.90\linewidth}{!}{
		\begin{tabular}{|c|c|c|} \hline
			\textbf{Input}                                                                                    & $x_t$ $(224\times 224\times3)$                                                      & $s_s$ $(224\times 224\times 3)$ \\
			\hline
			\multirow{3}{*}{\begin{tabular}[c]{@{}c@{}}\textbf{Encoder}
				                \end{tabular}}
			                                                                                                  & \multicolumn{2}{c|}{$\text{Conv}(F=64, K=7, S=1), \text{GroupNorm}, \text{GeLU}$}                                     \\
			                                                                                                  & \multicolumn{2}{c|}{$\text{Conv}(F=128, K=3, S=2), \text{GroupNorm}, \text{GeLU}$ }                                   \\
			                                                                                                  & \multicolumn{2}{c|}{$\text{Conv}(F=256, K=3, S=2), \text{GroupNorm}, \text{GeLU}$ }                                   \\
			\hline
			\multirow{3}{*}{\begin{tabular}[c]{@{}c@{}}\textbf{Intermediate} \\ \textbf{Blocks}\end{tabular}} & $z_t^G$ $(56\times56\times256)$
			                                                                                                  & $z_s^G$ $(56\times56\times256)$                                                                                       \\
			\cline{2-3}
			                                                                                                  & \multicolumn{2}{c|}{$z_s^{FIM} = \text{FIM}(z_t^G,z_s^G)$}                                                            \\
			\cline{2-3}
			                                                                                                  & \multicolumn{2}{c|}{$z_s^{FIM}$ $(56\times 56\times256)$}                                                             \\
			\hline
			\multirow{3}{*}{\begin{tabular}[c]{@{}c@{}}\textbf{Decoder}\end{tabular}}
			                                                                                                  & \multicolumn{2}{c|}{$\text{ConvT}(F=256, K=3, S=2), \text{GroupNorm}, \text{GeLU}$}                                   \\
			                                                                                                  & \multicolumn{2}{c|}{$\text{ConvT}(F=128, K=3, S=2), \text{GroupNorm}, \text{GeLU}$}                                   \\
			                                                                                                  & \multicolumn{2}{c|}{$\text{Conv}(F=64, K=7, S=1), \text{Tanh}$}                                                       \\
			\hline
			\textbf{Output}
			                                                                                                  & \multicolumn{2}{c|}{$\delta$ $(224\times224\times3)$}                                                                 \\
			\hline
		\end{tabular}
	}
	\label{tab:implementation_details}
\end{table}

\textbf{Training Details.} We show the detailed architecture of CD-MTA in \cref{tab:implementation_details}, with the shape of the tensors represented in the format of \texttt{[height, width, channels]}. The input shape differs for each dataset, \textit{e.g.}, \texttt{[224, 224, 3]} for ImageNet and \texttt{[32, 32, 3]} for CIFAR10~\cite{dataset_CIFAR}. Apart from the FIM, our encoder-decoder architecture follows a ResNet-like design which is similar to other generator-based methods. We used AdamW optimizer with a learning rate of $1e{-4}$ and a batch size of 12. We performed training on three different white-box classifiers: VGG19~\cite{cls_VGG}, ResNet-50 (Res50)~\cite{cls_ResNet}, and DenseNet-121 (Dense121)~\cite{cls_DenseNet}. For the loss in \cref{eq:feature_loss}, we used the feature maps of the last layer with a spatial resolution of $14 \times 14$ as this resolution yields the best results. This is further detailed in our ablation studies. The perturbation budget $\epsilon$ was set to 16/255, and we empirically selected $\lambda_{feat}=10$ as the best performing value. The generator was trained for 50 epochs on a single NVIDIA GTX 3090 GPU.

\textbf{Black-Box Multi-Targeted Attacks.} We used the 800 unseen ImageNet classes to evaluate the effectiveness of CD-MTA against other black-box multi-targeted attacks using the same evaluation protocol proposed by GAKer. We evaluated the attacks across nine classifiers with variying architectures: VGG19~\cite{cls_VGG}, Res50, ResNet-152 (Res152)~\cite{cls_ResNet}, Dense121~\cite{cls_DenseNet}, EfficientNet (EffNet)~\cite{cls_EffNet}, GoogLeNet~\cite{cls_GoogLeNet}, Inception-v3 (IncV3)~\cite{cls_inception}, ViT~\cite{cls_ViT}, and DeiT~\cite{cls_DeiT}.

\begin{table*}[!t]
	\centering
	\caption{
		\textbf{Black-Box Multi-targeted Attacks.} The multi-targeted attack success rate (\%) is averaged over the 800 unseen ImageNet classes. The top row indicates the model under attack (* indicates white-box attack). The top-right column shows the average over all models. Best results are in bold.
	}
	\setlength{\tabcolsep}{3pt}
	\resizebox{1.0\linewidth}{!}{
		\begin{tabular}{c|c|ccccccccc|c}
			\toprule
			\centering \textbf{Source} & \textbf{Attack}               & \textbf{Dense121}  & \textbf{Res50}     & \textbf{Res152}   & \textbf{VGG19}     & \textbf{EffNet}   & \textbf{GoogLeNet} & \textbf{IncV3}    & \textbf{ViT}      & \textbf{DeiT}     & \textbf{Avg.}     \\
			\midrule
			\multirow{5}{*}{{\textbf{VGG19}}}
			                           & C-GSP \cite{AdvAtt_C-GSP}     & 0.10               & 0.08               & 0.10              & 0.08*              & 0.06              & 0.06               & 0.04              & 0.10              & 0.07              & 0.08              \\
			                           & CGNC \cite{AdvAtt_CGNC}       & 1.68               & 1.22               & 0.84              & 3.70*              & 0.96              & 0.60               & 0.23              & 0.21              & 0.18              & 1.07              \\
			                           & Dual-Flow \cite{adv_dualflow} & 7.71               & 7.41               & 6.37              & 8.17*              & 5.27              & 6.17               & 3.48              & 1.80              & 1.17              & 5.28              \\
			                           & GAKer \cite{AdvAtt_GAKer}     & 13.33              & 11.05              & 5.41              & 43.25*             & 15.00             & 3.40               & 3.20              & 2.28              & 10.20             & 11.90             \\
			                           & \CC CD-MTA (Ours)             & \CC\textbf{20.09}  & \CC\textbf{30.03}  & \CC\textbf{30.32} & \CC\textbf{62.94*} & \CC\textbf{41.47} & \CC\textbf{8.87}   & \CC\textbf{4.12}  & \CC\textbf{5.68}  & \CC\textbf{12.13} & \CC\textbf{23.96} \\
			\midrule
			\multirow{5}{*}{{\textbf{Res50}}}
			                           & C-GSP \cite{AdvAtt_C-GSP}     & 0.06               & 0.05*              & 0.14              & 0.15               & 0.04              & 0.06               & 0.04              & 0.05              & 0.06              & 0.07              \\
			                           & CGNC \cite{AdvAtt_CGNC}       & 2.09               & 2.45*              & 1.66              & 1.84               & 1.03              & 1.14               & 0.42              & 0.34              & 0.23              & 1.24              \\
			                           & Dual-Flow \cite{adv_dualflow} & 11.90              & 11.92*             & 10.33             & 11.06              & 9.54              & 10.18              & 5.71              & 3.75              & 2.36              & 8.53              \\
			                           & GAKer \cite{AdvAtt_GAKer}     & 23.80              & 41.69*             & 23.05             & 26.02              & 13.40             & 11.49              & 5.85              & 1.44              & 4.99              & 16.86             \\
			                           & \CC CD-MTA (Ours)             & \CC\textbf{34.64}  & \CC\textbf{63.31*} & \CC\textbf{56.90} & \CC\textbf{33.40}  & \CC\textbf{48.17} & \CC\textbf{18.86}  & \CC\textbf{7.04}  & \CC\textbf{8.87}  & \CC\textbf{15.09} & \CC\textbf{31.81} \\
			\midrule
			\multirow{5}{*}{{\textbf{Dense121}}}
			                           & C-GSP \cite{AdvAtt_C-GSP}     & 0.05*              & 0.06               & 0.05              & 0.09               & 0.02              & 0.04               & 0.00              & 0.03              & 0.01              & 0.04              \\
			                           & CGNC \cite{AdvAtt_CGNC}       & 2.80*              & 1.61               & 1.38              & 1.82               & 1.00              & 1.14               & 0.42              & 0.35              & 0.25              & 1.20              \\
			                           & Dual-Flow \cite{adv_dualflow} & 12.56*             & 10.29              & 9.37              & 10.12              & 8.68              & 10.73              & 7.15              & 3.60              & 1.70              & 8.24              \\
			                           & GAKer \cite{AdvAtt_GAKer}     & 40.31*             & 23.10              & 17.28             & 25.17              & 13.08             & 14.63              & 7.23              & 2.60              & 7.56              & 16.77             \\
			                           & \CC CD-MTA (Ours)             & \CC\textbf{52.65*} & \CC\textbf{44.11}  & \CC\textbf{48.88} & \CC\textbf{29.33}  & \CC\textbf{44.01} & \CC\textbf{26.48}  & \CC\textbf{11.42} & \CC\textbf{10.78} & \CC\textbf{13.29} & \CC\textbf{31.22} \\
			\bottomrule
		\end{tabular}
	}
	\label{tab:results_imagenet_unknown}
\end{table*}

\begin{figure*}[!t]
	\centering
	\begin{subfigure}{0.32\linewidth}
		\includegraphics[width=\linewidth]{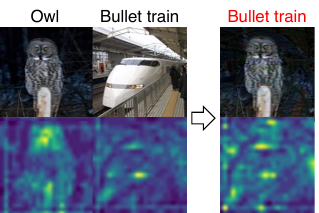}
		\caption{Target: Bullet train}
		\label{fig:res_1}
	\end{subfigure}
	\hfill
	\rule[.48cm]{0.1mm}{3.01cm} 
	\hfill
	\begin{subfigure}{0.32\linewidth}
		\includegraphics[width=\linewidth]{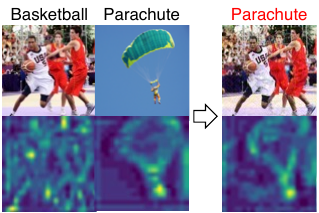}
		\caption{Target: Parachute}
		\label{fig:res_2}
	\end{subfigure}
	\hfill
	\rule[.48cm]{0.1mm}{3.01cm} 
	\hfill
	\begin{subfigure}{0.32\linewidth}
		\includegraphics[width=\linewidth]{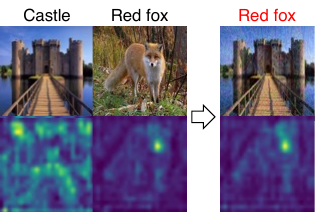}
		\caption{Target: Red fox}
		\label{fig:res_3}
	\end{subfigure}
	\caption{
		\textbf{Visualization of adversarial examples generated by CD-MTA using unseen ImageNet classes.} Each subfigure shows the source, target and pertubed images (top row), with their corresponding feature maps extracted from VGG19 and averaged across channels (bottom row). The perturbed images are aligned with the target images in the feature space, demonstrating the ability of CD-MTA to generate targeted adversarial examples without relying on class semantics.
	}
	\label{fig:res_vis_feats}
\end{figure*}

\textbf{Cross-Domain Multi-Targeted Attacks.} We assessed the feasibility of cross-domain multi-targeted attacks by evaluating the performance of our model on other datasets. Our benchmark used the same datasets as untargeted cross-domain settings~\cite{AdvAtt_BIA,yangFACLAttackFrequencyAwareContrastive2024,liUCGUniversalCrossDomain2024}. Specifically, we used three fine-grained datasets: CUB-200-2011~\cite{dataset_CUB}, Stanford Cars~\cite{dataset_STCar} and FGVC Aircraft~\cite{dataset_AIR}, and four coarse-grained datasets: CIFAR-10, CIFAR-100~\cite{dataset_CIFAR}, STL-10~\cite{dataset_STL10}, and SVHN~\cite{dataset_SVHN}. For fine-grained datasets, we tested three different black-box classifiers: Res50, SENet154, and SE-ResNet101~\cite{cls_SENet}. For coarse-grained dataset, we used one classifier per dataset from a publicly available repository\footnotemark[1]. It is important to note that the black-box classifiers were pretrained on their respective datasets, while the attack models were trained on ImageNet using a white-box classifier also pretrained on ImageNet. This setup ensures that the attacks are evaluated in a true cross-domain scenario, where they cannot benefit from data leakage. During inference, the source and target images were from different classes to avoid trivial solutions.

\textbf{Baselines.} We compared CD-MTA against C-GSP~\cite{AdvAtt_C-GSP}, CGNC~\cite{AdvAtt_CGNC}, Dual-Flow~\cite{adv_dualflow}, and GAKer~\cite{AdvAtt_GAKer}. Other methods are excluded from our evaluation, as they are inherently incapable to perform multi-targeted attacks. For cross-domain evaluation, C-GSP is omitted because it uses class labels as conditional inputs, making it inapplicable to unknown datasets which have different label sets. For CGNC~\cite{AdvAtt_CGNC} and Dual-Flow~\cite{adv_dualflow}, we follow the original training protocol, where prompts of the form “a photo of a \{\textit{class}\}” are passed to the CLIP text encoder~\cite{CLIP}. For Dual-Flow~\cite{adv_dualflow}, we also used the same pretrained Stable Diffusion model~\cite{StableDiffusion} suggested by the original authors. The only modification introduced for fairness is the use of the 200 ImageNet class subset proposed by GAKer to ensures that all baselines are trained and evaluated with the same data splits.

\textbf{Metric.} We evaluated the attack performance using the multi-targeted attack success rate (\%), which measures the proportion of perturbed images classified as the designated target class. All the reported success rates were averaged across the total number of classes.

\footnotetext[1]{\url{https://github.com/aaron-xichen/pytorch-playground}}

\subsection{Results on Black-Box Multi-Targeted Attack}

\begin{table*}[!t] \centering
	\caption{
		\textbf{Cross-Domain Multi-targeted Attacks on fine-grained datasets.} The attack was trained on ImageNet while the victim models were trained on fine-grained datasets. We report the multi-targeted attack success rate (\%) averaged over all categories of the corresponding dataset (indicated in the top row). The top-right column shows the average over all datasets and models. Best results are in bold.
	}
	\setlength{\tabcolsep}{3pt}
	\resizebox{1.0\linewidth}{!}{
		\begin{tabular}{c|c|ccc|ccc|ccc|c}
			\multicolumn{1}{c}{}       & \multicolumn{1}{c}{}          & \multicolumn{3}{c}{\textbf{CUB-200-2011}} & \multicolumn{3}{c}{\textbf{FGVC Aircraft}} & \multicolumn{3}{c|}{\textbf{Stanford Cars}} & \textbf{ALL}                                                                                                                                \\
			\toprule
			\centering \textbf{Source} & \textbf{Attack}               & \textbf{Res50}                            & \textbf{SENet154}                          & \textbf{SE-Res101}                          & \textbf{Res50}    & \textbf{SENet154} & \textbf{SE-Res101} & \textbf{Res50}    & \textbf{SENet154} & \textbf{SE-Res101} & \textbf{Avg.}     \\
			\midrule
			\multirow{4}{*}{{\textbf{VGG19}}}
			                           & CGNC \cite{AdvAtt_CGNC}       & 0.08                                      & 0.10                                       & 0.10                                        & 0.08              & 0.06              & 0.09               & 0.05              & 0.08              & 0.11               & 0.08              \\
			                           & Dual-Flow \cite{adv_dualflow} & 4.83                                      & 3.26                                       & 3.20                                        & 2.09              & 1.17              & 1.30               & 1.28              & 0.79              & 0.93               & 2.09              \\
			                           & GAKer \cite{AdvAtt_GAKer}     & 0.45                                      & 1.03                                       & 0.49                                        & 0.57              & 0.94              & 0.66               & 0.23              & 0.38              & 0.16               & 0.55              \\
			                           & \CC CD-MTA (Ours)             & \CC\textbf{6.26}                          & \CC\textbf{10.77}                          & \CC\textbf{5.91}                            & \CC\textbf{6.40}  & \CC\textbf{16.63} & \CC\textbf{5.27}   & \CC\textbf{4.83}  & \CC\textbf{14.66} & \CC\textbf{6.46}   & \CC\textbf{8.58}  \\
			\midrule
			\multirow{4}{*}{{\textbf{Res50}}}
			                           & CGNC \cite{AdvAtt_CGNC}       & 0.09                                      & 0.09                                       & 0.10                                        & 0.09              & 0.08              & 0.15               & 0.04              & 0.06              & 0.04               & 0.08              \\
			                           & Dual-Flow \cite{adv_dualflow} & 7.11                                      & 5.85                                       & 5.01                                        & 2.01              & 1.08              & 0.89               & 2.52              & 1.28              & 1.44               & 3.13              \\
			                           & GAKer \cite{AdvAtt_GAKer}     & 1.14                                      & 0.93                                       & 0.86                                        & 1.18              & 0.69              & 0.74               & 0.48              & 0.26              & 0.19               & 0.72              \\
			                           & \CC CD-MTA (Ours)             & \CC\textbf{9.07}                          & \CC\textbf{10.68}                          & \CC\textbf{8.37}                            & \CC\textbf{13.37} & \CC\textbf{21.11} & \CC\textbf{10.45}  & \CC\textbf{8.87}  & \CC\textbf{17.12} & \CC\textbf{10.36}  & \CC\textbf{12.16} \\
			\midrule
			\multirow{4}{*}{{\textbf{Dense121}}}
			                           & CGNC \cite{AdvAtt_CGNC}       & 0.08                                      & 0.09                                       & 0.08                                        & 0.11              & 0.05              & 0.10               & 0.03              & 0.09              & 0.04               & 0.07              \\
			                           & Dual-Flow \cite{adv_dualflow} & 7.69                                      & 5.89                                       & 5.31                                        & 2.06              & 1.32              & 1.29               & 2.98              & 1.56              & 1.69               & 3.42              \\
			                           & GAKer \cite{AdvAtt_GAKer}     & 1.16                                      & 1.15                                       & 1.55                                        & 0.96              & 0.86              & 0.95               & 0.42              & 0.35              & 0.44               & 0.87              \\
			                           & \CC CD-MTA (Ours)             & \CC\textbf{12.84}                         & \CC\textbf{12.39}                          & \CC\textbf{10.75}                           & \CC\textbf{16.30} & \CC\textbf{20.16} & \CC\textbf{12.98}  & \CC\textbf{12.39} & \CC\textbf{18.73} & \CC\textbf{15.59}  & \CC\textbf{14.68} \\
			\bottomrule
		\end{tabular}
	}
	\label{tab:results_fine_grained}
\end{table*}
\begin{table}[!t]
	\centering
	\caption{
		\textbf{Cross-Domain Multi-targeted Attacks on coarse-grained datasets.} The attack was trained on ImageNet while the victim models were trained on coarse-grained datasets. We report the multi-targeted attack success rate (\%) averaged over all categories of the corresponding dataset (indicated in the top row). The top-right column shows the average over all datasets and models. Best results are in bold.
	}
	\resizebox{1.0\linewidth}{!}{
		\begin{tabular}{c|c|cccc|c}
			\toprule
			\centering \textbf{Source} & \textbf{Attack}               & \textbf{cifar10}  & \textbf{cifar100} & \textbf{stl10}   & \textbf{svhn}     & \textbf{Avg.}     \\
			\midrule
			\multirow{4}{*}{{\textbf{VGG19}}}
			                           & CGNC \cite{AdvAtt_CGNC}       & 0.89              & 0.31              & 2.92             & 0.73              & 1.21              \\
			                           & Dual-Flow \cite{adv_dualflow} & 3.01              & 0.77              & 5.51             & 2.08              & 2.84              \\
			                           & GAKer \cite{AdvAtt_GAKer}     & 4.60              & 0.94              & 3.92             & 2.10              & 2.89              \\
			                           & \CC CD-MTA (Ours)             & \CC\textbf{13.09} & \CC\textbf{4.26}  & \CC\textbf{5.80} & \CC\textbf{16.56} & \CC\textbf{9.93}  \\
			\midrule
			\multirow{4}{*}{{\textbf{Res50}}}
			                           & CGNC \cite{AdvAtt_CGNC}       & 0.92              & 0.34              & 2.81             & 0.67              & 1.19              \\
			                           & Dual-Flow \cite{adv_dualflow} & 2.87              & 0.62              & 6.47             & 1.84              & 2.95              \\
			                           & GAKer \cite{AdvAtt_GAKer}     & 3.60              & 0.82              & 3.90             & 2.00              & 2.58              \\
			                           & \CC CD-MTA (Ours)             & \CC\textbf{11.76} & \CC\textbf{4.02}  & \CC\textbf{6.69} & \CC\textbf{19.26} & \CC\textbf{10.43} \\
			\midrule
			\multirow{4}{*}{{\textbf{Dense121}}}
			                           & CGNC \cite{AdvAtt_CGNC}       & 0.87              & 0.30              & 2.65             & 0.70              & 1.13              \\
			                           & Dual-Flow \cite{adv_dualflow} & 2.71              & 0.68              & 6.54             & 2.45              & 3.10              \\
			                           & GAKer \cite{AdvAtt_GAKer}     & 5.00              & 0.79              & 4.80             & 2.80              & 3.35              \\
			                           & \CC CD-MTA (Ours)             & \CC\textbf{8.95}  & \CC\textbf{2.92}  & \CC\textbf{6.65} & \CC\textbf{15.72} & \CC\textbf{8.56}  \\
			\bottomrule
		\end{tabular}
	}
	\label{tab:results_coarse_grained}
\end{table}

We show the attack performance on the 800 unseen ImageNet classes in \cref{tab:results_imagenet_unknown}. Our CD-MTA consistently outperforms baselines. For instance, when using VGG19 as the source model, our attack against Res152 achieved 24.91 and 29.48 points higher than GAKer and CGNC, respectively.

These results are particularly significant as \text{CD-MTA} does not rely on any pretrained information about the target class. In contrast, CGNC and Dual-Flow benefit from the use of CLIP, which is pretrained on large-scale datasets~\cite{CLIP,DG_CLIP}. Dual-Flow further benefits from Stable Diffusion~\cite{StableDiffusion}, pretrained on broad image-text distributions. GAKer also has an advantage as it uses a white-box classifier pretrained on the full ImageNet dataset during inference, which means that the 800 unseen classes used for evaluation are implicitely known by the attacker. Despite their strong advantages, \text{CD-MTA} still achieves superior performance, demonstrating the effectiveness of our class-agnostic feature objectives.

The visualizations in \cref{fig:res_vis_feats} further illustrate the effectiveness of CD-MTA. The perturbations successfully replace the source image's features with those of the target image. This is consistent with our goal to generate adversarial examples that are spatially aligned with the source data in the visual space, while also being aligned with the target data in the feature space.

\subsection{Results in Cross-Domain Multi-Targeted Attack}

We present the cross-domain results in \cref{tab:results_fine_grained,tab:results_coarse_grained}. Although modest, the results for CD-MTA are significantly better than prior methods. For example, when using Res50 as the source model, our model achieved 21.11\% against SENet154 pretrained on FGVC Aircraft, while GAKer, Dual-Flow, and CGNC only reached 0.69\%, 1.08\% and 0.28\%, respectively. This validates the feasibility of cross-domain targeted attacks, which were previously only possible in untargeted setting. These performance can further be improved by ensemble training, as discussed in \cref{sec:ensemble_training}.

We attribute the cross-domain transferability of our model to its ability to localize important features from the target input by leveraging spatial information. The effectiveness of CD-MTA on both fine-grained (\cref{tab:results_fine_grained}) and coarse-grained datasets (\cref{tab:results_coarse_grained}) demonstrates that the attack is capable to capture subtle patterns from unfamiliar domains. In contrast, existing methods achieve near 0\% success rate in most cases, indicating their inability to extract any useful information from the conditional input.

To perform targeted attacks, our method requires a single reference image of the desired target class at inference time. Some may interpret this as a form of data leakage. However, our setting is considerably more realistic and constrained than previous methods, which rely on full access to the target dataset during training. In practice, an attacker aiming for a specific target prediction is expected to know what the target class visually represents and can reasonably prepare or obtain a single image that the victim model would likely classify into that class. In our experiments, the reference images are randomly selected from the evaluation dataset and are never used during the training of either the perturbation generator or the victim model. This ensures that our evaluation setup remains fair and faithful to a practical black-box attack scenario. Our setting also aligns with the “unseen” multi-targeted attack protocol introduced by GAKer, which we extend to the cross-domain setting.

\begin{table}[!t]
	\centering
	\caption{
		\textbf{Multi-targeted Attacks against defense methods.} The multi-targeted attack success rates (\%) are averaged over 800 unseen ImageNet classes. The top row indicates the defense method. Best results are in bold.
	}
	\resizebox{1.0\linewidth}{!}{
		\renewcommand{\arraystretch}{1.12} 
		\begin{tabular}{c|c|cccccc|c}
			\toprule
			\centering & \textbf{Attack}               & \textbf{SIN}        & \textbf{SIN-IN}    & \textbf{SIN\(_{\text{fine}}\)} & \textbf{Augmix}    & \(\mathbf{l_2}\)   & \(\mathbf{l_\infty}\) & \textbf{Avg.}     \\
			\midrule
			\multirow{5}{*}{{\rotatebox[origin=c]{90}{\textbf{VGG19}}}}
			           & C-GSP \cite{AdvAtt_C-GSP}     & 0.06                & 0.05               & 0.05                           & 0.06               & 0.04               & 0.05                  & 0.05              \\
			           & CGNC \cite{AdvAtt_CGNC}       & 0.23                & 0.96               & 1.20                           & 0.59               & 0.09               & 0.07                  & 0.52              \\
			           & Dual-Flow \cite{adv_dualflow} & 5.46                & 7.02               & 6.81                           & 6.66               & 2.60               & 0.11                  & 4.78              \\
			           & GAKer \cite{AdvAtt_GAKer}     & 0.94                & 6.50               & 10.60                          & 4.67               & 0.16               & 0.01                  & 3.81              \\
			           & \CC CD-MTA (Ours)             & \CC \textbf{6.59}   & \CC \textbf{14.80} & \CC \textbf{18.09}             & \CC \textbf{13.49} & \CC \textbf{3.12}  & \CC \textbf{0.15}     & \CC\textbf{9.37}  \\
			\midrule
			\multirow{5}{*}{{\rotatebox[origin=c]{90}{\textbf{Res50}}}}
			           & C-GSP \cite{AdvAtt_C-GSP}     & 0.05                & 0.04               & 0.04                           & 0.06               & 0.03               & 0.04                  & 0.04              \\
			           & CGNC \cite{AdvAtt_CGNC}       & 0.48                & 10.4               & 2.19                           & 1.36               & 0.09               & 0.13                  & 2.44              \\
			           & Dual-Flow \cite{adv_dualflow} & 9.93                & 11.63              & 11.24                          & 11.36              & 6.88               & 0.18                  & 8.54              \\
			           & GAKer \cite{AdvAtt_GAKer}     & 4.84                & 27.44              & 30.88                          & 18.18              & 0.11               & 0.08                  & 13.59             \\
			           & \CC CD-MTA (Ours)             & \CC \textbf{14.18}  & \CC \textbf{29.15} & \CC \textbf{33.12}             & \CC \textbf{29.96} & \CC \textbf{8.27}  & \CC \textbf{0.30}     & \CC\textbf{19.16} \\
			\midrule
			\multirow{5}{*}{{\rotatebox[origin=c]{90}{\textbf{Dense121}}}}
			           & C-GSP \cite{AdvAtt_C-GSP}     & 0.05                & 0.05               & 0.04                           & 0.05               & 0.04               & 0.06                  & 0.05              \\
			           & CGNC \cite{AdvAtt_CGNC}       & 0.52                & 1.54               & 1.59                           & 1.06               & 0.12               & 0.10                  & 0.82              \\
			           & Dual-Flow \cite{adv_dualflow} & 10.58               & 10.88              & 9.61                           & 10.58              & 6.84               & 0.33                  & 8.14              \\
			           & GAKer \cite{AdvAtt_GAKer}     & 4.12                & 19.35              & 21.01                          & 15.20              & 0.13               & 0.00                  & 9.97              \\
			           & \CC CD-MTA (Ours)             & \CC  \textbf{21.91} & \CC \textbf{35.51} & \CC \textbf{33.91}             & \CC \textbf{29.78} & \CC \textbf{13.72} & \CC \textbf{0.56}     & \CC\textbf{22.57} \\
			\bottomrule
		\end{tabular}
	}
	\label{tab:results_defense}
\end{table}

\begin{table}[!t]
	\centering
	\caption{
		\textbf{Untargeted Attacks against defense methods.} We use untargeted attack success rate (\%), which is the percentage of images misclassified by the defense model. The top row indicates the defense method. Best results are in bold. The second best are underlined.
	}
	\resizebox{1.0\linewidth}{!}{
		\renewcommand{\arraystretch}{1.12} 
		\begin{tabular}{c|c|cccccc|c}
			\toprule
			\centering & \textbf{Attack}               & \textbf{SIN}          & \textbf{SIN-IN}       & \textbf{SIN\(_{\text{fine}}\)} & \textbf{Augmix}       & \(\mathbf{l_2}\)      & \(\mathbf{l_\infty}\)  & \textbf{Avg.}        \\
			\midrule
			\multirow{5}{*}{{\rotatebox[origin=c]{90}{\textbf{VGG19}}}}
			           & C-GSP \cite{AdvAtt_C-GSP}     & 67.54                 & 72.24                 & 76.85                          & 66.27                 & 56.81                 & 37.56                  & 62.88                \\
			           & CGNC \cite{AdvAtt_CGNC}       & 69.51                 & 69.51                 & 78.56                          & 68.46                 & 59.66                 & 39.38                  & 64.18                \\
			           & Dual-Flow \cite{adv_dualflow} & \textbf{84.83}        & \textbf{83.81}        & \textbf{83.49}                 & \textbf{79.97}        & \textbf{75.09}        & \textbf{43.39}         & \textbf{75.10}       \\
			           & GAKer \cite{AdvAtt_GAKer}     & 70.12                 & 76.18                 & 80.29                          & 72.82                 & 59.71                 & 39.53                  & 66.44                \\
			           & \CC CD-MTA (Ours)             & \CC \underline{70.33} & \CC \underline{76.85} & \CC \underline{81.21}          & \CC \underline{73.96} & \CC \underline{62.31} & \CC  \underline{39.61} & \CC\underline{67.38} \\
			\midrule
			\multirow{5}{*}{{\rotatebox[origin=c]{90}{\textbf{Res50}}}}
			           & C-GSP \cite{AdvAtt_C-GSP}     & 80.13                 & 87.62                 & 89.69                          & 79.58                 & 72.33                 & 41.44                  & 75.13                \\
			           & CGNC \cite{AdvAtt_CGNC}       & 78.42                 & 89.20                 & 91.56                          & 79.47                 & 73.15                 & 41.15                  & 75.49                \\
			           & Dual-Flow \cite{adv_dualflow} & \textbf{85.98}        & 83.81                 & 83.49                          & 81.04                 & 71.62                 & 41.31                  & 74.54                \\
			           & GAKer \cite{AdvAtt_GAKer}     & \underline{85.17}     & \textbf{95.16}        & \textbf{97.10}                 & \underline{88.67}     & \underline{72.74}     & \underline{41.47}      & \textbf{80.05}       \\
			           & \CC CD-MTA (Ours)             & \CC 83.42             & \CC \underline{91.21} & \CC \underline{92.91}          & \CC \textbf{89.17}    & \CC \textbf{77.23}    & \CC \textbf{42.65}     & \CC\underline{79.43} \\
			\midrule
			\multirow{5}{*}{{\rotatebox[origin=c]{90}{\textbf{Dense121}}}}
			           & C-GSP \cite{AdvAtt_C-GSP}     & 77.57                 & 82.61                 & 84.02                          & 75.54                 & 70.34                 & 41.44                  & 71.92                \\
			           & CGNC \cite{AdvAtt_CGNC}       & 79.15                 & 86.56                 & 88.24                          & 79.78                 & 75.80                 & 43.21                  & 75.46                \\
			           & Dual-Flow \cite{adv_dualflow} & \underline{86.17}     & 84.66                 & 84.44                          & 82.78                 & 76.42                 & 44.64                  & 76.52                \\
			           & GAKer \cite{AdvAtt_GAKer}     & 85.19                 & \underline{91.21}     & \underline{92.05}              & \underline{85.79}     & \underline{76.76}     & \underline{45.79}      & \underline{79.47}    \\
			           & \CC CD-MTA (Ours)             & \CC \textbf{89.29}    & \CC \textbf{94.24}    & \CC \textbf{94.54}             & \CC \textbf{90.15}    & \CC \textbf{83.39}    & \CC \textbf{46.61}     & \CC\textbf{83.04}    \\
			\bottomrule
		\end{tabular}
	}
	\label{tab:results_defense_un}
\end{table}

\subsection{Robustness Against Defenses}

We conducted additional experiments against robust models, including adversarially trained Res50 ($l_2$ and $l_\infty$)~\cite{AdvAtt_advRobustImageNet}, Stylized ImageNet (SIN, SIN-IN, SIN\(_{\text{fine}}\))~\cite{SL_texture_bias_Geirhos}, and Augmix~\cite{hendrycksAugMixSimpleData2020}. While the performance decreases in \cref{tab:results_defense}, CD-MTA is still able to achieve multi-targeted attacks with better success rates than other methods. Additionally, the untargeted attack results in \cref{tab:results_defense_un} show that the drop in performance in multi-targeted settings does not imply a complete failure of the attack. CD-MTA maintains high untargeted success rates, posing a non-negligible level of threat against robust models. Although Dual-Flow occasionally yields higher untargeted success rates, this is likely due to its reliance on Stable Diffusion~\cite{StableDiffusion}. Despite this, CD-MTA still achieves superior attack than other baselines in most cases.

It is also worth noting that model owners may assume that restricting access to the architecture, weights, and training data ensures the security of their model against adversarial attacks. This could lead to weaker defense measures because of the inherent trade-off between model robustness and accuracy~\cite{Adv_PIAT,DBLP:conf/iclr/RadeM22,raghunathanUnderstandingMitigatingTradeoff2020}. Our work highlights this overlooked threat and raises awareness of the necessity for strong security measures even when the training data is kept private.

\begin{figure}[!t] \centering
	\includegraphics[width=1.00\linewidth]{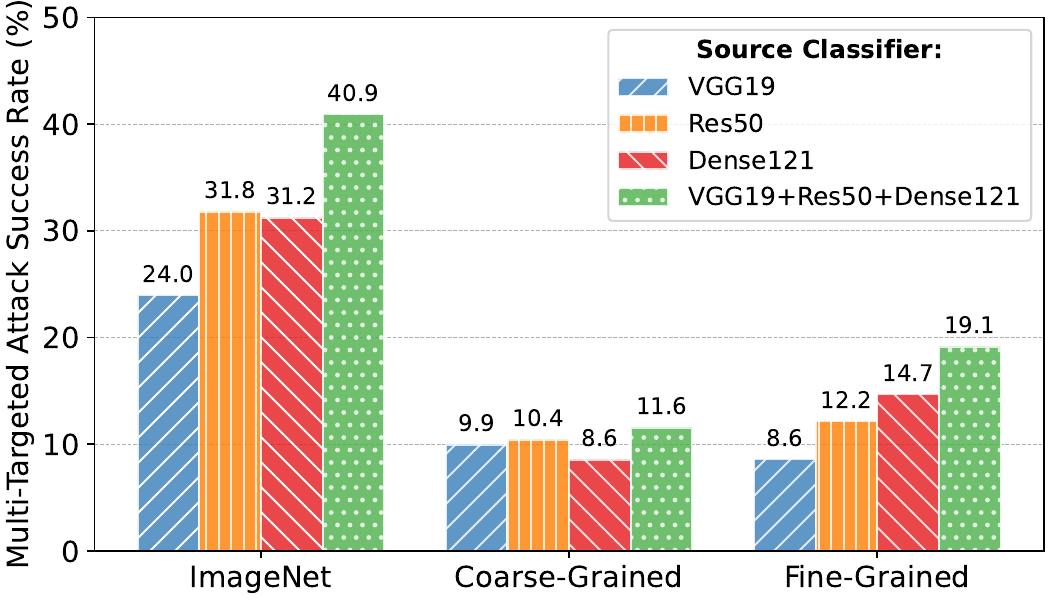}
	\caption{
		Multi-targeted attack success rates (\%) of CD-MTA w/ and w/o ensemble training, evaluated on unseen ImageNet classes, coarse-grained datasets, and fine-grained datasets. The results are averaged across all categories and classifiers.
	}
	\label{fig:ensemble_results}
\end{figure}

\subsection{Ensemble Training}\label{sec:ensemble_training}

Ensemble training is a common technique to improve the performance of adversarial attacks~\cite{chenAdaptiveModelEnsemble2023}. It consists on training the perturbation generator with multiple white-box classifiers to enhance its generalization capabilities. Given $n$ classifiers $f_1,\ldots, f_n$, we denote $f_{i,l_i}(\cdot)$ as the feature maps extracted by the $i$-th classifier at layer $l_i$. The feature loss $L_{feat}$ in \cref{eq:feature_loss} can then be replaced by \cref{eq:ensemble_feature_loss} for ensemble training:

\begin{align}\label{eq:ensemble_feature_loss}
	L_{feat} & = 1 - \sum_{i=1}^n \cos(f_{i,l_i}(x_s'), f_{i,l_i}(x_t)).
\end{align}

\begin{figure*}[!t]
	\centering
	\begin{subfigure}{0.32\linewidth}
		\includegraphics[width=\linewidth]{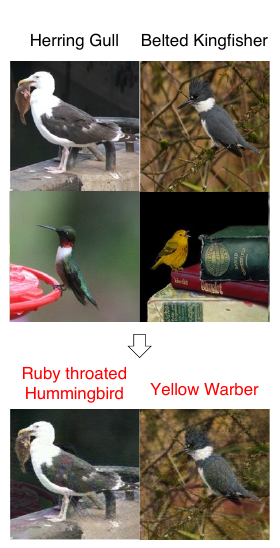}
		\caption{CUB-200-2011}
		\label{fig:viz_CUB}
	\end{subfigure}
	\hfill
	\rule[.58cm]{0.1mm}{8.13cm} 
	\hfill
	\begin{subfigure}{0.32\linewidth}
		\includegraphics[width=\linewidth]{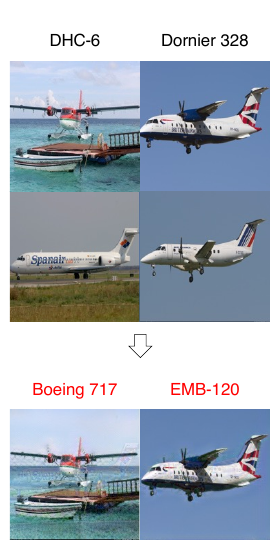}
		\caption{FGVC Aircraft}
		\label{fig:viz_AIR}
	\end{subfigure}
	\hfill
	\rule[.58cm]{0.1mm}{8.13cm} 
	\hfill
	\begin{subfigure}{0.32\linewidth}
		\includegraphics[width=\linewidth]{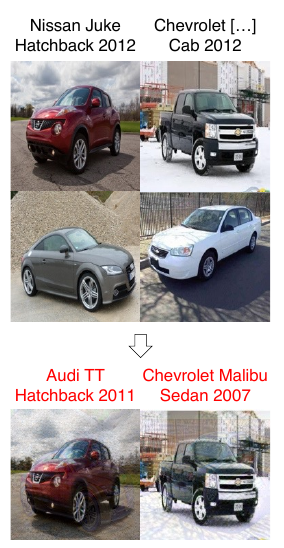}
		\caption{Stanford Cars}
		\label{fig:viz_CAR}
	\end{subfigure}
	\centering
	\begin{subfigure}{0.23\linewidth}
		\includegraphics[width=\linewidth]{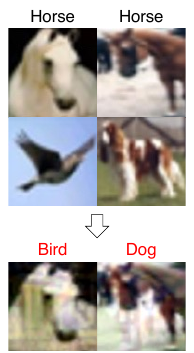}
		\caption{cifar10}
		\label{fig:viz_cifar10}
	\end{subfigure}
	\hfill
	\rule[.45cm]{0.1mm}{5.95cm} 
	\hfill
	\begin{subfigure}{0.23\linewidth}
		\includegraphics[width=\linewidth]{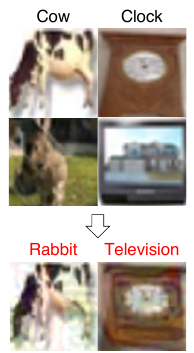}
		\caption{cifar100}
		\label{fig:viz_cifar100}
	\end{subfigure}
	\hfill
	\rule[.45cm]{0.1mm}{5.95cm} 
	\hfill
	\begin{subfigure}{0.23\linewidth}
		\includegraphics[width=\linewidth]{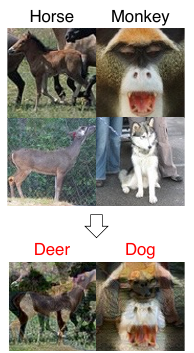}
		\caption{stl10}
		\label{fig:viz_stl10}
	\end{subfigure}
	\hfill
	\rule[.45cm]{0.1mm}{5.95cm} 
	\hfill
	\begin{subfigure}{0.23\linewidth}
		\includegraphics[width=\linewidth]{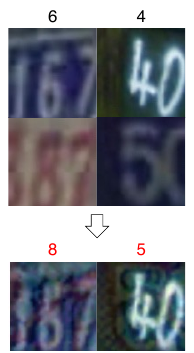}
		\caption{svhn}
		\label{fig:viz_svhn}
	\end{subfigure}
	\caption{
		Visualization of perturbed images generated by CD-MTA on fine-grained (\cref{fig:viz_CUB,fig:viz_AIR,fig:viz_CAR}) and coarse-grained (\cref{fig:viz_cifar10,fig:viz_cifar100,fig:viz_stl10,fig:viz_svhn}) datasets. The top row displays the clean source images with their ground-truth labels, while the middle row presents the target images. The bottom row shows the perturbed images produced by CD-MTA, with their predicted labels matching with the ground-truth target labels. All the noises are restricted by the $L_\infty$-norm with $\epsilon=16/255$.\\
	}
	\label{fig:viz_cross_domain}
\end{figure*}

\begin{table*}[!t]
	\centering
	\caption{
		Loss ablation study on the 800 unseen ImageNet classes in multi-targeted attack (* denotes white-box attack).
	}
	\resizebox{1.0\linewidth}{!}{
		\begin{tabular}{c|cc|ccccccccc|c}
			\toprule
			\centering \textbf{Source} & \textbf{$L_{feat}$} & \textbf{$L_{fr}$} & \textbf{Dense121} & \textbf{Res50}     & \textbf{Res152}   & \textbf{VGG19}    & \textbf{EffNet}   & \textbf{GoogLeNet} & \textbf{IncV3}    & \textbf{ViT}      & \textbf{DeiT}     & \textbf{Avg.}     \\
			\midrule
			\multirow{2}{*}{{\textbf{VGG19}}}
			                           & ---                 & \checkmark        & 0.04              & 0.03               & 0.13              & 0.19*             & 0.12              & 0.04               & 0.06              & 0.39              & 0.21              & 0.13              \\
			                           & \checkmark          & ---               & 17.67             & 28.21              & 27.19             & \textbf{65.81*}   & 38.95             & 7.39               & 3.38              & 4.09              & 11.55             & 22.69             \\
			                           & \CC\checkmark       & \CC\checkmark     & \CC\textbf{20.09} & \CC\textbf{30.03}  & \CC\textbf{30.32} & \CC62.94*         & \CC\textbf{41.47} & \CC\textbf{8.87}   & \CC\textbf{4.12}  & \CC\textbf{5.68}  & \CC\textbf{12.13} & \CC\textbf{23.96} \\
			\midrule
			\multirow{3}{*}{{\textbf{Res50}}}
			                           & ---                 & \checkmark        & 0.07              & 0.10*              & 0.18              & 0.07              & 0.19              & 0.11               & 0.10              & 0.13              & 0.72              & 0.19              \\
			                           & \checkmark          & ---               & 33.92             & \textbf{63.31*}    & 56.27             & 33.38             & 46.12             & 18.21              & 6.23              & 4.08              & 7.89              & 29.93             \\
			                           & \CC\checkmark       & \CC\checkmark     & \CC\textbf{34.64} & \CC\textbf{63.31*} & \CC\textbf{56.90} & \CC\textbf{33.40} & \CC\textbf{48.17} & \CC\textbf{18.86}  & \CC\textbf{7.04}  & \CC\textbf{8.87}  & \CC\textbf{15.09} & \CC\textbf{31.81} \\
			\midrule
			\multirow{2}{*}{{\textbf{Dense121}}}
			                           & ---                 & \checkmark        & 0.08*             & 0.06               & 0.17              & 0.11              & 0.17              & 0.13               & 0.08              & 0.09              & 0.68              & 0.17              \\
			                           & \checkmark          & ---               & \textbf{53.13*}   & 44.06              & 48.25             & \textbf{29.37}    & 44.01             & 26.34              & 11.13             & 10.17             & 13.17             & 31.07             \\
			                           & \CC\checkmark       & \CC\checkmark     & \CC52.65*         & \CC\textbf{44.11}  & \CC\textbf{48.88} & \CC29.33          & \CC\textbf{44.14} & \CC\textbf{26.48}  & \CC\textbf{11.42} & \CC\textbf{10.78} & \CC\textbf{13.29} & \CC\textbf{31.23} \\
			\bottomrule
		\end{tabular}
	}
	\label{tab:ablation_imagenet_unknown}
\end{table*}
\begin{table*}[!t]
	\centering
	\caption{
		Loss ablation study on fine-grained datasets.
	}
	\resizebox{1.0\linewidth}{!}{
		\begin{tabular}{c|cc|ccc|ccc|ccc|c}
			\multicolumn{1}{c}{}       & \multicolumn{2}{c}{} & \multicolumn{3}{c}{\textbf{CUB-200-2011}} & \multicolumn{3}{c}{\textbf{FGVC Aircraft}} & \multicolumn{3}{c|}{\textbf{Stanford Cars}} & \textbf{ALL}                                                                                                                                                     \\
			\toprule
			\centering \textbf{Source} & \textbf{$L_{feat}$}  & \textbf{$L_{fr}$}                         & \textbf{Res50}                             & \textbf{SENet154}                           & \textbf{SE-Res101} & \textbf{Res-50}   & \textbf{SENet154} & \textbf{SE-Res101} & \textbf{Res-50}   & \textbf{SENet154} & \textbf{SE-Res101} & \textbf{Avg.}     \\
			\midrule
			\multirow{3}{*}{{\textbf{VGG19}}}
			                           & ---                  & \checkmark                                & 0.08                                       & 0.20                                        & 0.12               & 0.18              & 0.21              & 0.18               & 0.03              & 0.02              & 0.0                & 0.11              \\
			                           & \checkmark           & ---                                       & 5.39                                       & 9.07                                        & 4.92               & 4.76              & 14.62             & 3.84               & 3.52              & 11.39             & 4.44               & 6.88              \\
			                           & \CC\checkmark        & \CC\checkmark                             & \CC\textbf{6.26}                           & \CC\textbf{10.77}                           & \CC\textbf{5.91}   & \CC\textbf{6.40}  & \CC\textbf{16.63} & \CC\textbf{5.27}   & \CC\textbf{4.83}  & \CC\textbf{14.66} & \CC\textbf{6.46}   & \CC\textbf{8.58}  \\
			\midrule
			\multirow{3}{*}{{\textbf{Res50}}}
			                           & ---                  & \checkmark                                & 0.13                                       & 0.14                                        & 0.17               & 0.17              & 0.22              & 0.22               & 0.06              & 0.09              & 0.06               & 0.14              \\
			                           & \checkmark           & ---                                       & 8.36                                       & 9.75                                        & 7.67               & 12.29             & 20.12             & 9.60               & 8.12              & 15.94             & 10.09              & 11.33             \\
			                           & \CC\checkmark        & \CC\checkmark                             & \CC\textbf{9.07}                           & \CC\textbf{10.68}                           & \CC\textbf{8.37}   & \CC\textbf{13.37} & \CC\textbf{21.11} & \CC\textbf{10.45}  & \CC\textbf{8.87}  & \CC\textbf{17.12} & \CC\textbf{10.36}  & \CC\textbf{12.16} \\
			\midrule
			\multirow{3}{*}{{\textbf{Dense121}}}
			                           & ---                  & \checkmark                                & 0.11                                       & 0.15                                        & 0.09               & 0.20              & 0.17              & 0.19               & 0.04              & 0.09              & 0.08               & 0.12              \\
			                           & \checkmark           & ---                                       & 12.71                                      & 12.37                                       & 10.66              & 15.44             & 19.50             & 12.38              & 4.37              & 13.51             & 6.08               & 11.89             \\
			                           & \CC\checkmark        & \CC\checkmark                             & \CC\textbf{12.84}                          & \CC\textbf{12.46}                           & \CC\textbf{10.75}  & \CC\textbf{16.30} & \CC\textbf{20.16} & \CC\textbf{12.99}  & \CC\textbf{12.39} & \CC\textbf{18.73} & \CC\textbf{15.59}  & \CC\textbf{14.69} \\
			\bottomrule
		\end{tabular}
	}
	\label{tab:ablation_fine_grained}
\end{table*}
\begin{table}[!t]
	\centering
	\caption{
		Loss ablation study on coarse-grained datasets.
	}
	\resizebox{1.0\linewidth}{!}{
		\begin{tabular}{c|cc|cccc|c}
			\toprule
			\centering \textbf{Source} & \textbf{$L_{feat}$} & \textbf{$L_{fr}$} & \textbf{cifar10}  & \textbf{cifar100} & \textbf{stl10}   & \textbf{svhn}     & \textbf{Avg.}     \\
			\midrule
			\multirow{3}{*}{{\textbf{VGG19}}}
			                           & ---                 & \checkmark        & 3.12              & 0.89              & 3.61             & 6.04              & 3.42              \\
			                           & \checkmark          & ---               & 12.62             & 4.14              & 5.34             & 14.56             & 9.17              \\
			                           & \CC\checkmark       & \CC\checkmark     & \CC\textbf{13.09} & \CC\textbf{4.26}  & \CC\textbf{5.80} & \CC\textbf{16.56} & \CC\textbf{9.93}  \\
			\midrule
			\multirow{3}{*}{{\textbf{Res50}}}
			                           & ---                 & \checkmark        & 3.03              & 0.83              & 4.50             & 6.89              & 3.81              \\
			                           & \checkmark          & ---               & 11.55             & 3.99              & 6.62             & 19.14             & 10.33             \\
			                           & \CC\checkmark       & \CC\checkmark     & \CC\textbf{11.76} & \CC\textbf{4.02}  & \CC\textbf{6.69} & \CC\textbf{19.26} & \CC\textbf{10.43} \\
			\midrule
			\multirow{3}{*}{{\textbf{Dense121}}}
			                           & ---                 & \checkmark        & 1.67              & 0.71              & 4.38             & 6.16              & 3.23              \\
			                           & \checkmark          & ---               & \textbf{8.95}     & 2.83              & \textbf{6.66}    & 14.84             & 8.32              \\
			                           & \CC\checkmark       & \CC\checkmark     & \CC\textbf{8.95}  & \CC\textbf{2.92}  & \CC6.65          & \CC\textbf{15.72} & \CC\textbf{8.56}  \\
			\bottomrule
		\end{tabular}
	}
	\label{tab:ablation_coarse_grained}
\end{table}

We evaluated the impact of ensemble training for using three white-box classifiers: VGG19, Dense121, and Res50. We evaluated the attack performance on unseen ImageNet classes, coarse-grained datasets, and fine-grained datasets. We compared the results with the same model trained with a single classifier. The results in \cref{fig:ensemble_results} show that ensemble training significantly improves the attack performance. The multi-targeted attack success rate on unseen ImageNet classes increased by an average of 10\% when using 3 white-box classifiers, and by 5\% on fine-grained datasets.


\subsection{Detailed Qualitative Results}

We provide additional qualitative results of our method in \cref{fig:viz_cross_domain}. The targeted attack proves effective across a wide range of target classes, including those that are never seen during training. In particular, the generator successfully adapts to visually diverse categories and produces semantically meaningful perturbations that consistently lead the classifier to the desired target class. These examples further illustrate the generalization ability of our approach in cross-domain, black-box scenarios.

\subsection{Ablation Studies}

The loss ablation study in \cref{tab:ablation_imagenet_unknown,tab:ablation_coarse_grained,tab:ablation_fine_grained} shows that $L_{fr}$ improves the attack against black-box models, while slightly reducing performance against white-box models, with up to 9.51\% gain against SE-Res101 on Stanford Cars. The performance drop for white-box models aligns with the idea that the FRO encourages the perturbation generator to be less dependent on the white-box model. The results also indicate that the attack fails without $L_{feat}$, suggesting that the generator still requires guidance from a substitute model for effective noise optimization.

\begin{figure}[!t]
	\centering
	\includegraphics[width=1.00\linewidth]{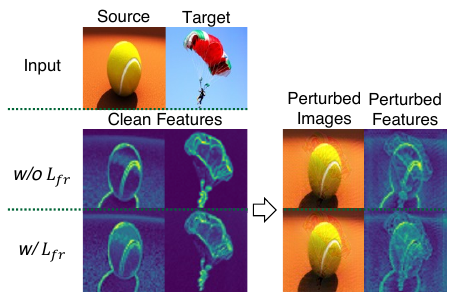}
	\caption{
		Feature maps extracted from the perturbation generator w/ and w/o $L_{fr}$. The two perturbed features look significantly different, while the clean features are almost identical.
	}
	\label{fig:ablation_features}
\end{figure}

\begin{figure}[!t]
	\centering
	\begin{subfigure}{\linewidth}
		\centering
		\includegraphics[width=0.99\linewidth]{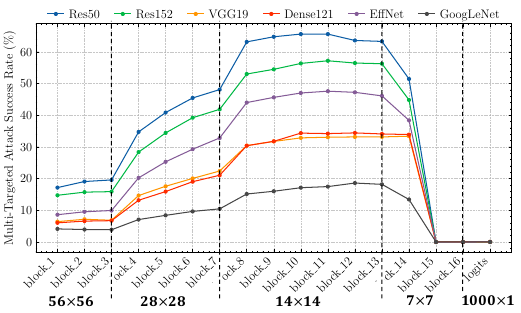}
		\vspace{-0.45em}
		\caption{ImageNet unseen classes (Source Domain).}
		\vspace{0.55em}
		\label{fig:layers_imagenet}
	\end{subfigure}
	\begin{subfigure}{\linewidth}
		\centering
		\includegraphics[width=0.99\linewidth]{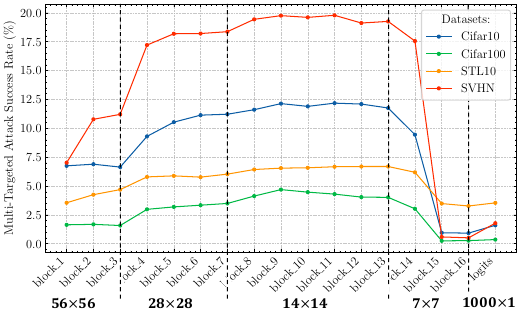}
		\vspace{-0.45em}
		\caption{Coarse-grained datasets (Cross-Domain).}
		\vspace{0.55em}
		\label{fig:layers_coarse}
	\end{subfigure}
	\begin{subfigure}{\linewidth}
		\centering
		\includegraphics[width=0.99\linewidth]{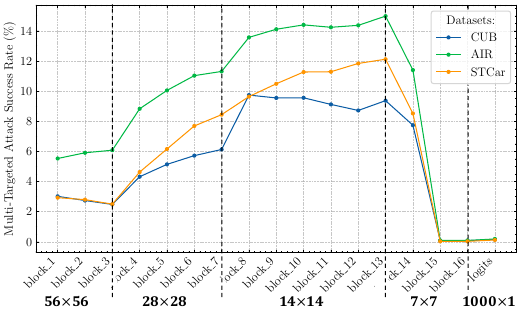}
		\vspace{-0.45em}
		\caption{Fine-grained datasets (Cross-Domain).}
		\label{fig:layers_fine}
	\end{subfigure}
	\caption{
		Multi-targeted attack success rates (\%) obtained by training CD-MTA on different blocks of Res50. Each \textit{block\_$l$} represents an attack trained using the $l$-th block of Res50. The corresponding spatial resolution $H\times W$ is indicated below the $x$-axis. The results are averaged over all classes.
	}
	\label{fig:layers_all}
\end{figure}

We additionally visualized the feature maps extracted from the perturbation generator with and without $L_{fr}$ in \cref{fig:ablation_features}. The results reveal that $L_{fr}$ decreases the intensity of the source information in the perturbed features while slightly enhancing the presence of the target information. This observation is consistent with our expectation that the FRO encourages the generator to preserve target-relevant information in the generated perturbations.

To determine the optimal feature extraction layer for $L_{feat}$, we conducted ablation studies by applying the attack to different layers of the white-box classifier. We trained CD-MTA on each block of Res50 and evaluated its transferability to black-box models in \cref{fig:layers_imagenet,fig:layers_coarse,fig:layers_fine}. The results indicate that feature maps with a resolution of $14 \times 14$ are the most effective across all datasets, with no significant difference between blocks at this scale. Shallower layers likely lack sufficient discriminative features, as the white-box model has not yet effectively processed the input. Conversely, the low performance observed when using logits ($K = 1000$) or $7 \times 7$ feature maps suggests that deeper layers capture strong dataset-specific and class-specific biases, which hinder generalization to unseen classes. This is particularly relevant given that GAKer relies on a logits-based objective; our results indicate that such losses are not suitable for cross-domain generalization. This observation demonstrates the necessity of our $L_{feat}$ design instead of using logits-based losses commonly employed in prior multi-targeted attacks.

\begin{figure}[!t]
	\centering
	\includegraphics[width=1.00\linewidth]{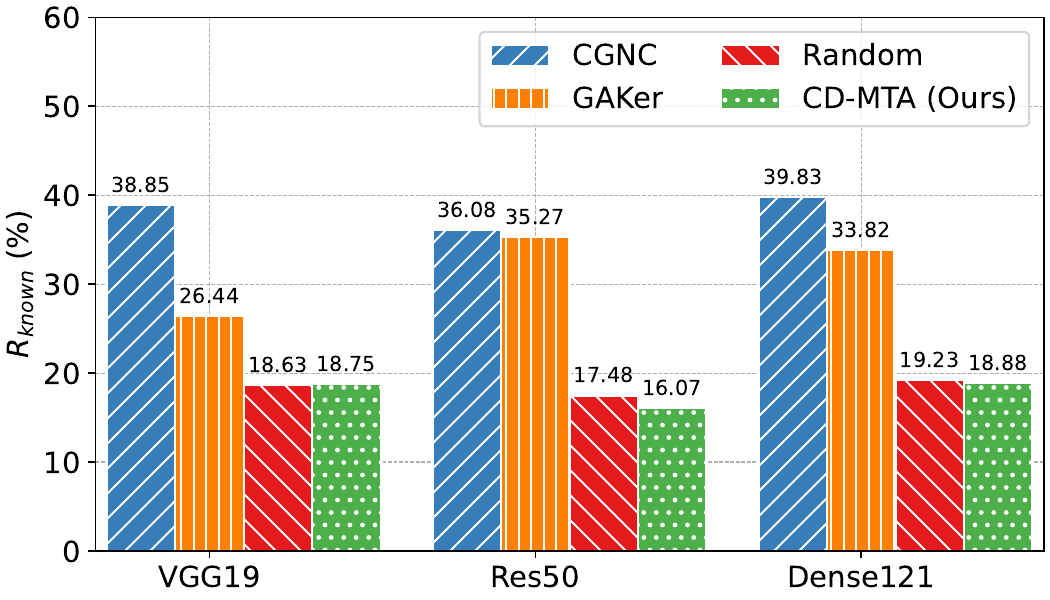}
	\caption{
		Proportion $R_{known}$ of perturbed images misclassified to one of the 200 classes seen during training. The perturbations were generated using only the 800 unseen classes.
	}
	\label{fig:exp_overfit}
\end{figure}

\subsection{Identifying Overfitting in Multi-Targeted Attacks}

CGNC demonstrated high performance on known classes as reported in~\cite{AdvAtt_CGNC}, but failed to target unseen classes as shown in~\cref{tab:results_imagenet_unknown}. This is unexpected since its use of CLIP should enhance generalizability~\cite{CLIP,DG_CLIP}. These findings suggest that CGNC, along with other targeted attack methods, may suffer from overfitting to the classes learned during training.

To test this hypothesis, we generated adversarial examples using only the 800 unseen classes and measured how often they were misclassified to one of the 200 known classes. For each attack methods (CGNC, GAKer and \text{CD-MTA}), we generated 40k adversarial examples exclusively from the unseen classes. These examples were then fed to the black-box classifiers, and we retained only those that were misclassified. From this subset, we computed the proportion $R_{known}$ of images redirected into one of the 200 known classes. As a non-overfitting baseline, we repeated the experiment using random noise perturbations.

The results in \cref{fig:exp_overfit} show that CGNC and GAKer redirected approximately 30\% of perturbed images into one of the 200 known classes, while random noises resulted in a ratio of around 20\%. This clearly indicates overfitting to learned semantics, as they ignored the original conditional input and instead generated perturbations toward one of the known classes. In contrast, CD-MTA achieved a ratio comparable to the random baseline, demonstrating its ability to prevent overfitting and confirming its class-agnostic nature.

\section{Societal Impacts \& Limitations}

CD-MTA reveals a significant blind spot in the current understanding of black-box model security. In real-world scenarios, many models have hidden architectures, weights, and training data, making them more difficult to attack. Existing cross-domain targeted attacks implicitely assume access to the training data, which is often not the case in practice. Our work demonstrates that targeted attacks remain feasible even when the class sets of the victim model are unknown.

Despite its strengths, CD-MTA faces limitations against robust models. Although it performs better than existing methods, its success rate declines under strong defenses. This suggests that while CD-MTA generalizes well across classes and datasets, it remains vulnerable to modern defense techniques. This highlight the importance of developing robust defenses for DNNs, even when the model is completely private.

\section{Conclusion}

In this paper, we identified the problem of data leakage in existing cross-domain targeted attacks and introduced CD-MTA, a model capable of performing such attacks without relying on the victim model's training data. The key idea is to transform a target sample into perturbations without inferring its class semantics, by extracting fine-grained spatial information using class-agnostic objective functions. This enables generalization to unseen classes and datasets, while mitigating the overfitting to known classes observed in existing methods. Extensive experiments show that CD-MTA outperforms state-of-the-art methods by over 20\% in multi-targeted settings and up to 17.26\% in cross-domain targeted attacks. These results demonstrate the feasibility of such attacks even when training data remains private, exposing potential vulnerabilities in neural networks that are assumed to be secure dues to the confidentiality of their training data. Our findings underscore the importance of developing robust defenses against such threats. Future work may explore improving cross-domain generalization and adapting these ideas to tasks beyond classification, such as object detection and segmentation.

{
    \small
    \bibliographystyle{ieeenat_fullname}
    \bibliography{main}
}

\end{document}